%% file: sigconf.tex
\documentclass[sigconf]{acmart}

\usepackage[utf8]{inputenc}
\usepackage{afterpage}
\usepackage{booktabs}
\usepackage{float}
\usepackage{multicol}
\usepackage{multirow}
\usepackage{subcaption}
\usepackage{./slashbox}
\usepackage[labelfont=bf, font=small]{caption}

\copyrightyear{2018} 
\acmYear{2018} 
\setcopyright{acmcopyright}
\acmConference[SAC 2018]{SAC 2018: Symposium on Applied Computing }{April 9--13, 2018}{Pau, France}
\acmPrice{15.00}
\acmDOI{10.1145/3167132.3167237}
\acmISBN{978-1-4503-5191-1/18/04}

\begin{document}
\title[A smartphone application to measure the quality of spraying machines via image analysis]
    {A smartphone application to measure the quality of pest control spraying machines via image analysis}

\author{Bruno~Brandoli~Machado}
\affiliation{%
    \centering
    Computer Science Department\\
    Federal University of Mato~Grosso~do~Sul\\
    Ponta Pora, Brazil
}
\authornote{Corresponding author --- \url{brunobrandoli@gmail.com}.}

\author{Gabriel~Spadon}
\affiliation{%
    \centering
    Institute of Mathematics and~Computer Science\\
    University of Sao Paulo\\
    Sao Carlos, Brazil
}

\author{Mauro~S.~Arruda}
\affiliation{%
    \centering
    Computer Science Department\\
    Federal University of Mato~Grosso~do~Sul\\
    Ponta Pora, Brazil
}

\author{Wesley~N.~Goncalves}
\affiliation{%
    \centering
    Computer Science Department\\
    Federal University of Mato~Grosso~do~Sul\\
    Ponta Pora, Brazil
}

\author{Andre~C.~P.~L.~F.~Carvalho}
\affiliation{%
    \centering
    Institute of Mathematics and~Computer Science\\
    University of Sao Paulo\\
    Sao Carlos, Brazil
}

\author{Jose~F.~Rodrigues-Jr}
\affiliation{%
    \centering
    Institute of Mathematics and~Computer Science\\
    University of Sao Paulo\\
    Sao Carlos, Brazil
}

\renewcommand{\shortauthors}{B. B. Machado et al.}

\begin{abstract}
The need for higher agricultural productivity has demanded the intensive use of pesticides.
However, their correct use depends on assessment methods that can accurately predict how well the pesticides' spraying covered the intended crop region.
Some methods have been proposed in the literature, but their high cost and low portability harm their widespread use.
This paper proposes and experimentally evaluates a new methodology based on the use of a smartphone-based mobile application, named DropLeaf.
Experiments performed using  DropLeaf showed that, in addition to its versatility, it can predict with high accuracy the pesticide spraying.
DropLeaf is a five-fold image-processing methodology based on: (i) color space conversion; (ii) threshold noise removal; (iii) convolutional operations of dilation and erosion; (iv) detection of contour markers in the water-sensitive card; and, (v) identification of droplets via the marker-controlled watershed transformation.
The authors performed successful experiments over two case studies, the first using a set of synthetic cards and the second using a real-world crop.
The proposed tool can be broadly used by farmers equipped with conventional mobile phones, improving the use of pesticides with health, environmental and financial benefits.
\end{abstract}

%
%

\ccsdesc[500]{Human-centered computing~Mobile computing}
\ccsdesc[500]{Human-centered computing~Mobile phones}
\ccsdesc[500]{Computing methodologies~Image segmentation}
\ccsdesc[300]{Computing methodologies~Computer vision}

\keywords{Mobile Application, Image Processing,\\ Pesticide Spraying Analysis, Deposition Analysis}

\pagenumbering{gobble}
\sloppy

\maketitle

\input{body-conf}

\bibliographystyle{ACM-Reference-Format}
\bibliography{bibliography} 

\end{document}

%% file: body-conf.tex
\section{Introduction}
\label{sec:intro}

The world population is estimated to be 7 billion people with a projection of increasing to 9.2 billion by 2050.
An increase that will demand nearly 70\% more food due to changes in dietary habits (more dairy and grains) in underdeveloped countries~\cite{FAO2009}.
To cope with such a challenge, it is mandatory to increase the productivity of existing land, which is achieved by means of less waste along the food chain, and by the use of pesticides.
Pesticides correspond to chemical preparations for destroying weed plants (via herbicides), fungal (via fungicides), or insects (via insecticides)~\cite{Bon2014}.
The use of pesticides is disseminated worldwide, accounting for a 40-billion-dollar annual budget~\cite{Popp2013} with tons of chemicals (roughly 2 kg per hectare~\cite{Liu2015}) being applied in all kinds of crops with the aim of increasing the production of food.
Current trends point that a large range of agricultural and horticultural systems are to face heavier pressures from pests, leading to a higher demand for pesticides.

In this scenario, it is important that the correct amount of pesticide is sprayed on the crop fields.
Too much and there might be residues in the produced food along with environmental contamination; too little and there might be regions of the crop that are not protected, reducing the productivity.
Besides, irregular spray coverage might cause pest and/or weed resistance or behavioral avoidance~\cite{PS3773, PS3330}.
In order to evaluate the pulverization, it is necessary to measure the spray coverage, that is, the proportional area covered by the pesticide formulation droplets (water carrier, active ingredients, and adjuvant).

The problem of measuring the spray coverage abridges to knowing how much pesticide was sprayed on each part of the crop field.
The standard manner to do that is to distribute oil or water-sensitive cards (WSC) over the soil; such cards are coated with a bromoethyl dye that turns blue in the presence of water~\cite{Giles2003}.
The problem, then, becomes assessing each card by counting the number of droplets per unit area, by drawing their size distribution, and by estimating the percentage of the card area that was covered; these measures allow one to estimate the volume of sprayed pesticide per unit area of the crop.
If done manually, this process is burdensome and imprecise.
This is where automated solutions become the first need, motivating a number of commercial solutions including the Swath Kit~\cite{Mierzejewski1991}, a pioneer computer-based process that uses image processing to analyze the water-sensitive cards; the USDA-ARS system~\cite{Hoffman2005}, a camera-based system that uses 1-$cm^2$ samples from the cards to form a pool of sensor data; the DropletScan~\cite{Wolf2003}, a flatbed scanner defined over a proprietary hardware; the DepositScan system, made of a laptop computer and a handheld business card scanner~\cite{Zhu2011}; and the AgroScan system\footnote{~\url{http://www.agrotec.etc.br/produtos/agroscan/}}, a batch-based outsource service that performs analyzes over collected cards.
All these systems, however, are troublesome to carry throughout the field, requiring the collection, scanning, and post-processing of the cards, a time-consuming and labor-intensive process.
An alternative is to use wired, or wireless, sensors~\cite{Crowe2005}; an expensive solution that demands constant maintenance.

Since there is a consensus with respect to the need of achieving a homogeneous spray coverage to gain productivity in agricultural and horticultural systems, there is room for research in innovative means of evaluating the spraying of pesticides.
Such means might benefit from the current commodity technology found in mobile cell phones, which carry computing resources powerful enough to perform a wide range of applications.
In the form of a cell phone application, it is possible to conceive a readily-available solution, portable up to the crop field, to aid farmers and agronomists in the task of measuring the spray coverage and, hence, in the decision-making process concerning where and how to pulverize.
This is the aim of the present study, in which we introduce DropLeaf, a cell phone application able to estimate the amount of pesticide sprayed on water-sensitive cards.
DropLeaf works on regular smartphones, what significantly simplifies the assessment of the pesticide application.
It uses the cell phone's camera to capture images of the spray cards, instantly producing estimates of the spray coverage by means of image processing techniques.

The remainder of the paper is structured as follows.
Section~\ref{sec:approach} describes the steps of the proposed approach to measure the quality of pest control spraying.
In addition, in this section, we describe the techniques implemented in the mobile application.
In Section~\ref{sec:res}, we show the results achieved by our application.
Section~\ref{sec:discussao} reviews major points related to our results.
Conclusions come in Section~\ref{sec:conclusao}.

\section{Methodology \& application}
\label{sec:approach}

In this section, we introduce our methodology, named DropLeaf, to estimate the pesticide spray coverage.
The aim of the technique is to measure the coverage area of water-sensitive spray cards, so to aid in the estimation of the crop pesticide coverage, as discussed in Section~\ref{sec:intro}.
DropLeaf is based on image processing techniques built up on a mobile application that is functional on commodity cell phones.
The software calculates measures from the drops observed on the spray cards, presenting statistics that enable the assessment of the spraying:

\begin{itemize}
   \item{{\bf Coverage Density (CD)}: given in percentage of covered area per area unit in $cm^2$;}
   \item{{\bf Volumetric Median Diameter (VMD)}: given by the 50th percentile $D_{V0.5}$ of the diameter distribution;}
   \item{{\bf Diameter Relative Span (DRS)}: given by $DRS=\frac{D_{V0.9}-D_{V0.1}}{D_{V0.5}}$, where $D_{V0.1}$ is the 10th percentile and $D_{V0.9}$ is the 90th percentile of the diameter distribution.}
\end{itemize}

The three measures are used to understand how much of the field was covered with pesticide and how well the pesticide was dispersed; the finer the diameters and the higher the coverage area, the better the dispersion.

In order to calculate those measures, it is necessary to determine the diameter (in micrometers) of each drop observed in a given card.
Manually, this is a laborious task that might take hours per card.
Instead, DropLeaf uses an intricate image processing method that saves time and provides superior precision when compared to manual inspection and to former techniques.

\begin{figure*}[!htb]
   \centering
   \includegraphics[width=\linewidth]{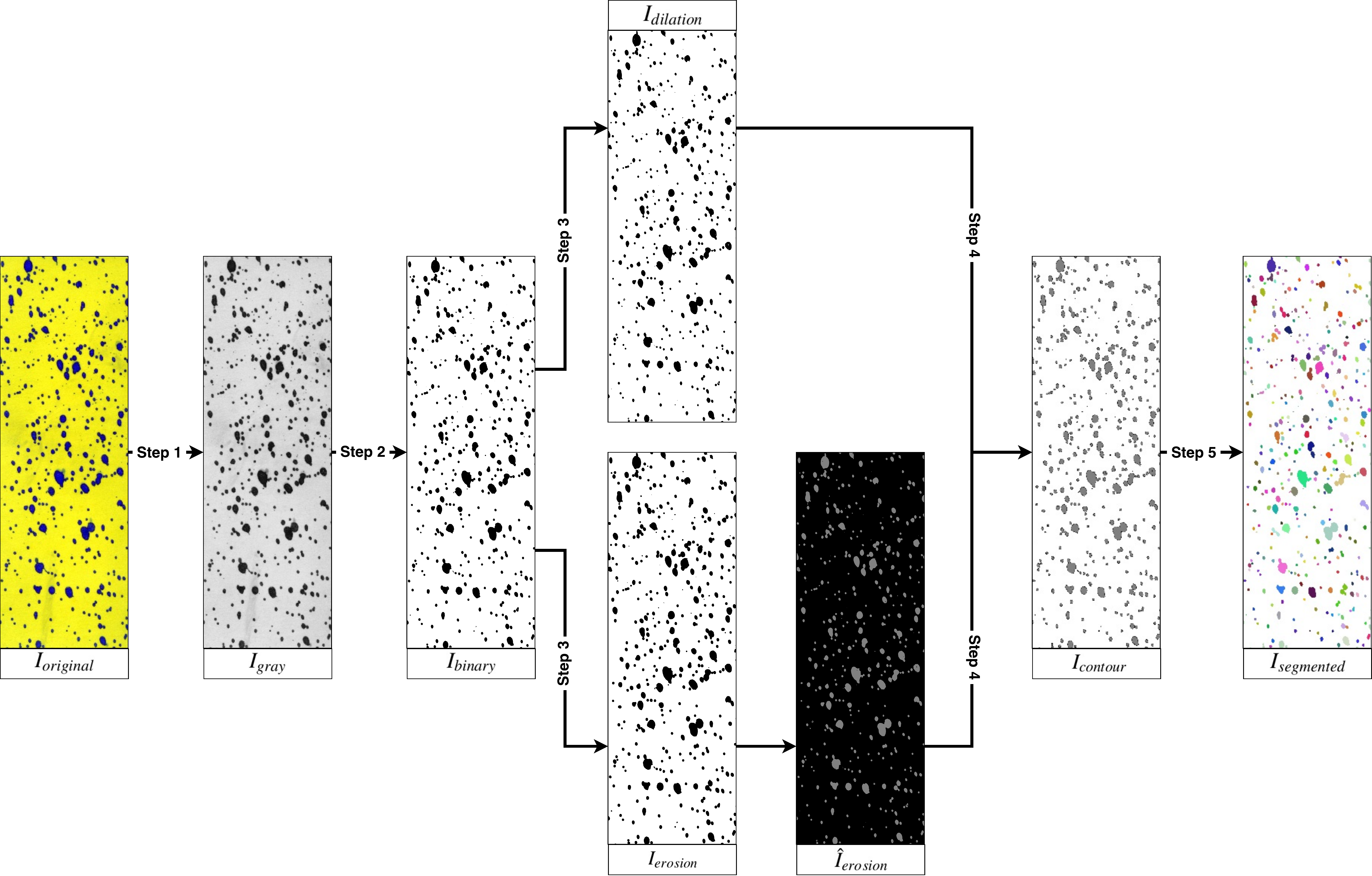}
   \caption
   {The image processing method of DropLeaf.
   It starts by loading an image of a water sensitive paper.
   Then, it performs a color-space transformation to obtain a grayscale version of the same image --- {\bf Step 1}.
   Subsequently, the grayscale image is binarized to isolate the drops and to remove noise --- {\bf Step 2}.
   Next, in two distinct copies of the noisy-removed image, it applies morphological operations of Dilation and Erosion; the dilated image is inverted to contrast with the color of the eroded one --- {\bf Step 3}.
   The next step is to compute the difference between the two images, delineating the contours (masks) of the droplets.
   The resulting image is used to identify the contours' markers that delimit the area of each drop --- {\bf Step 4}.
   Finally, the contours are used to segment the drops in sheds --- {\bf Step 5}, providing the tool with a well-defined set of droplets.}
   \label{fig:proposta}
\end{figure*}

Figure~\ref{fig:proposta} illustrates the image processing method of DropLeaf, which consists of five steps applied to each spray card image: (I) color space conversion to grayscale; (II) binarization by thresholding (noise removal); (III) dilation and erosion in distinct copies of the image; (IV) complement operation over dilated and eroded images to produce contour markers; (V) drop identification via marker-controlled watershed.
Following, we explain each step specifying why it was necessary and how it relates to the next step.
To illustrate the steps of the method, we provide a running example whose initial spray card image is presented in Figure~\ref{fig:proposta}(a).

\subsection{Grayscale transformation}

After the acquisition of an image via the cellphone camera $I_{original}(x,y) = (R_{xy}, G_{xy}, B_{xy}) \in [0,1]^3$, the Step 1 is to convert it to a grayscale image $I_{gray}(x,y) \in [0,1]$.
This is necessary to ease the discrimination of the card surface from the drops that fell on it.
We use the continuous domain of [0,1] so that our formalism is able to express any color depth; specifically we use 32 bits for RGB and 8 bits for grayscale.
Color information is not needed as it would make the computation heavier and more complex.
This first step, then, transforms the image into a grayscale representation, see Figure~\ref{fig:proposta}(b), according to:
\begin{equation}
   I_{gray}(x,y) = 0.299*R_{xy} + 0.587*G_{xy} + 0.114*B_{xy}
\end{equation}

\subsection{Binarization}

Here, the grayscale image $I_{gray}$ passes through a threshold-based binarization process -- Step 2, a usual step for image segmentation.
Since the grayscale is composed of a single color channel, binarization can be achieved simply by choosing a threshold value.
Gray values $I_{gray}(x,y)$ below the threshold become black, and $I_{gray}(x,y)$ values above the threshold become white.
Since spray cards are designed to stress the contrast between the card and the drops, the threshold value can be set as a constant value -- we use value $0.35$ corresponding to value $90$ in the 8-bit domain $[0,255]$.
This is a choice that removes noise and that favors faster processing if compared to more elaborated binarization processes like those based on clustering or on gray-levels distribution.
Figure~\ref{fig:proposta}(c) depicts the result, an image $I_{binary}(x,y) \in \{0,1\}$ given by:

\begin{equation}
   I_{binary}(x,y) =
       \begin{cases}
           0, & \mbox{if }I_{gray}(x,y) < 0.35\\
           1, & \mbox{otherwise}
       \end{cases}
\end{equation}

\subsection{Dilation and erosion}

At this point, we need to identify the contours of the drops -- Step 3, which will delimit their diameters.
We use an approach based on convolution operators of dilation $\oplus$ and erosion $\ominus$~\cite{gonzalez2007image}.
We proceed by creating two copies of the binary image.
One copy passes through dilation -- Figure~\ref{fig:proposta}(d), a morphological operation that probes and expands the shapes found in an image.
For dilation to occur, a structuring element (a square binary matrix) is necessary so to specify the extent of the dilation.
We used a $3 \times 3$ matrix $B$ so to dilate the drops by nearly 1 pixel.
Note that, at this point, we still do not know about the drops; rather, the dilation convolution has the mathematical property of interacting with potential shapes to be segmented, thus allowing for drop identification.
After dilation, the shapes that correspond to the drops will be 1 pixel larger all along their perimeters.
Formally, we produce an image $I_{dilation}$ according to:
\begin{equation}
   I_{dilation} = I_{binary} \oplus B
\end{equation}

After that, the second copy of the binary image passes through erosion -- Figure~\ref{fig:proposta}(e), also a morphological operation that, contrary to dilation, contracts the shapes found in the image.
Again, we use a $3 \times 3$ matrix $B$ as the structuring element so to erode the drops by nearly 1 pixel.
Formally, we produce an image $I_{erosion}$ according to:
\begin{equation}
   I_{erosion} = I_{binary} \ominus B
\end{equation}

\subsection{Contour identification}

Given the two images produced by dilation and erosion --- $I_{dilation}$ with drops larger than the original and $I_{erosion}$ with drops smaller than the original --- the trick then is to identify the contours of the drops by measuring the difference between the dilated and the eroded drops.
This is achieved by applying a complement operation over the two binary images -- Step 4.
To do so, first, we invert the eroded image, so that 1's become 0's and vice versa, obtaining image $\hat{I}_{erosion}=\neg I_{erosion}$.
Then, it is sufficient to perform the following pixel by pixel logic AND:
\begin{equation}
I_{contour}(x,y) = I_{dilation}(x,y)~\text{\bf AND}~\hat{I}_{erosion}(x,y)
\end{equation}

The result is the binary image $I_{contour}$ that is depicted in Figure~\ref{fig:proposta}(f), in which only contour pixels have value 1.

\subsection{Marker-based watershed segmentation}

In the last step -- Step 5, with contours properly marked on the image, we proceed to the drop identification considering the previously identified contours.
To this end, we used the marker-based watershed segmentation.
Watershed~\cite{vincent1991watersheds} is a technique that considers an image as a topographic relief in which the gray level of the pixels corresponds to their altitude.
The transform proceeds by simulating the flooding of the landscape starting at the local minima.
This process forms basins that are gradually fulfilled with water.
Eventually, the water from different basins meet, indicating the presence of ridges (boundaries); this is an indication that a segment was found and delimited.
The process ends when the water reaches the highest level in the color-encoding space.
The problem with the classical watershed is that it might over-segment the image in the case of an excessive number of minima.
For a better precision, we use the marker-controlled variation of the algorithm~\cite{gaetano2012marker}.
This variation is meant for images whose shapes define proper contours previously given to the algorithm.
Given the contours (markers), the marker-based watershed proceeds by considering as minima, only the pixels within the boundaries of the contours.
Watershed is an iterative algorithm computationally represented by a function {\it watershed(Image i, Image[] contours)}.
We use such a function to produce a set of segments (drops) over the gray-level image $I_{gray}$ while considering the set of contours identified in the image $I_{contour}$, as follows:
\begin{equation}
   watershed(I_{gray},findContours(I_{contour}))
\end{equation}

\noindent
where $Image[]~findContours(Image~i)$ is a function that, given an image, returns a set of sub images (matrices) corresponding to the contours found in the image; meanwhile, watershed is a function that, given an input image and a set of sub images corresponding to contours, produces a set of segments stored in the array of input contours.

We use the product of the watershed function to produce our final output $I_{segmented}$ simply by drawing the segments over the original image, as illustrated in Figure~\ref{fig:proposta}(g).
Notice, however, that the last image of the process, $I_{segmented}$, is meant only for visualization.
The analytical process, the core of the methodology, is computed over the set of segments.

\subsection{Diameter processing}

After segmentation is concluded, we have a set of segments, each corresponding to a drop of pesticide.
The final step is to compute the measures presented at the beginning of this section: coverage density (CD), volumetric median diameter (VMD), and diameter relative span (DRS).
Since we have the segments computationally represented by an array of binary matrices, we can calculate the area and the diameters of each drop by counting the pixels of each matrix.
After counting, it is necessary to convert the diameter given in pixels into a diameter given in micrometers ($\mu m$), which, for the i-th drop, goes as follows:
\begin{equation}
   diameter_{\mu m}(drop_i) = width_{px}(drop_i)*\frac{width^{card}_{\mu m}}{width^{card}_{px}}
   \label{eq:diametermu}
\end{equation}

\noindent
where, $width_{px}(drop_i)$ is the width in pixels of the $i$-th drop; $width^{card}_{px}$ is the width of the card in pixels; and $width^{card}_{\mu m}$ is the width of the card in micrometers.
Notice that we used $width$, but we could have used $height$ as well; what matters is that the fraction provides a conversion ratio given in $px/\mu m$, which is not sensible to the axis; horizontal or vertical, the ratio is the same for a non-distorted image.

Notice that $width_{px}(drop_i)$ and $width^{card}_{px}$ are obtainable via image processing, after the segmentation method; meanwhile, $width^{card}_{\mu m}$ is a constant provided by the user, corresponding to the real-world width of the card.
Also, notice that we are considering that the diameter corresponds to the horizontal axis (the width) of the drop; it is possible, however, that the diameter corresponds to the vertical axis, in which case the formulation is straightly similar.
Choosing between the horizontal and the vertical axes might be tricky in case the drop is elliptical, rather than circular.
We solved this issue by extracting the diameter from the area of the drop.
We use the formula of the circle area $a_{circle}=\pi*radius^2 = \pi*(\frac{diameter}{2})^2$.
With simple algebra, we conclude that given the area in pixels of the $i$-th drop, its diameter in pixels is given by the following equation:
\begin{equation}
   diameter_{px}(drop_i) = 2*\sqrt{\frac{area_{px}(drop_i)}{\pi}}
   \label{eq:diameterrefined}
\end{equation}

Rewriting Equation~\ref{eq:diametermu} by means of Equation~\ref{eq:diameterrefined}, we get:
\begin{equation}
   diameter_{\mu m}(drop_i) = 2*\sqrt{\frac{area_{px}(drop_i)}{\pi}}*\frac{width^{card}_{\mu m}}{width^{card}_{px}}
   \label{eq:diametermufinal}
\end{equation}

Once the diameter is converted into micrometers, it becomes trivial to compute all the measures that support the spray card analysis, as described in the beginning of Section~\ref{sec:approach}.

\subsection*{Implementation details}

The use of mobile devices to perform automatic tasks has increased fast~\cite{fengMS2015}.
The main reasons for it are the recent advances in hardware, such as sensors, processors, memories, and cameras.
Thereby, smartphones have become new platforms for applications of image processing and computer vision~\cite{farinellaPR2015}.

\begin{figure*}[!htb]
   \centering
   \begin{subfigure}{.32\textwidth}
       \centering
       \includegraphics[width=.8\linewidth]{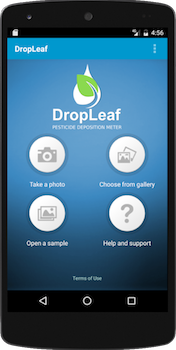}
   \end{subfigure}
   \begin{subfigure}{.32\textwidth}
       \centering
       \includegraphics[width=.8\linewidth]{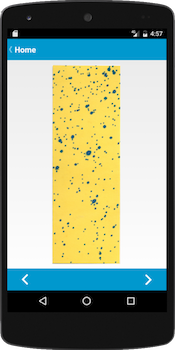}
   \end{subfigure}
   \begin{subfigure}{.32\textwidth}
       \centering
       \includegraphics[width=.8\linewidth]{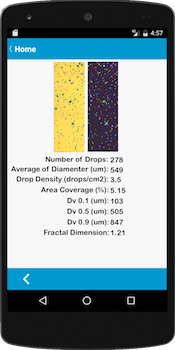}
   \end{subfigure}
    \caption{A preview of our fully functional application.}
    \label{fig:sample}
\end{figure*}

Mobile devices are an adequate mean to perform tasks in real-time {\it in situ} far from the laboratory.
In this context, besides this methodology, the contribution of this paper is the development of a mobile application to measure the quality of pesticide spraying on water-sensitive cards.
The methods that constitute the process were imported from the OpenCV library\footnote{~\url{http://opencv.org}}.
Java was the programming language.
The application is fully functional as depicted in Figure~\ref{fig:sample}.

\begin{figure}[!b]
   \centering
   \setlength{\fboxsep}{0pt}
   \fbox{\includegraphics[width=.70\linewidth]{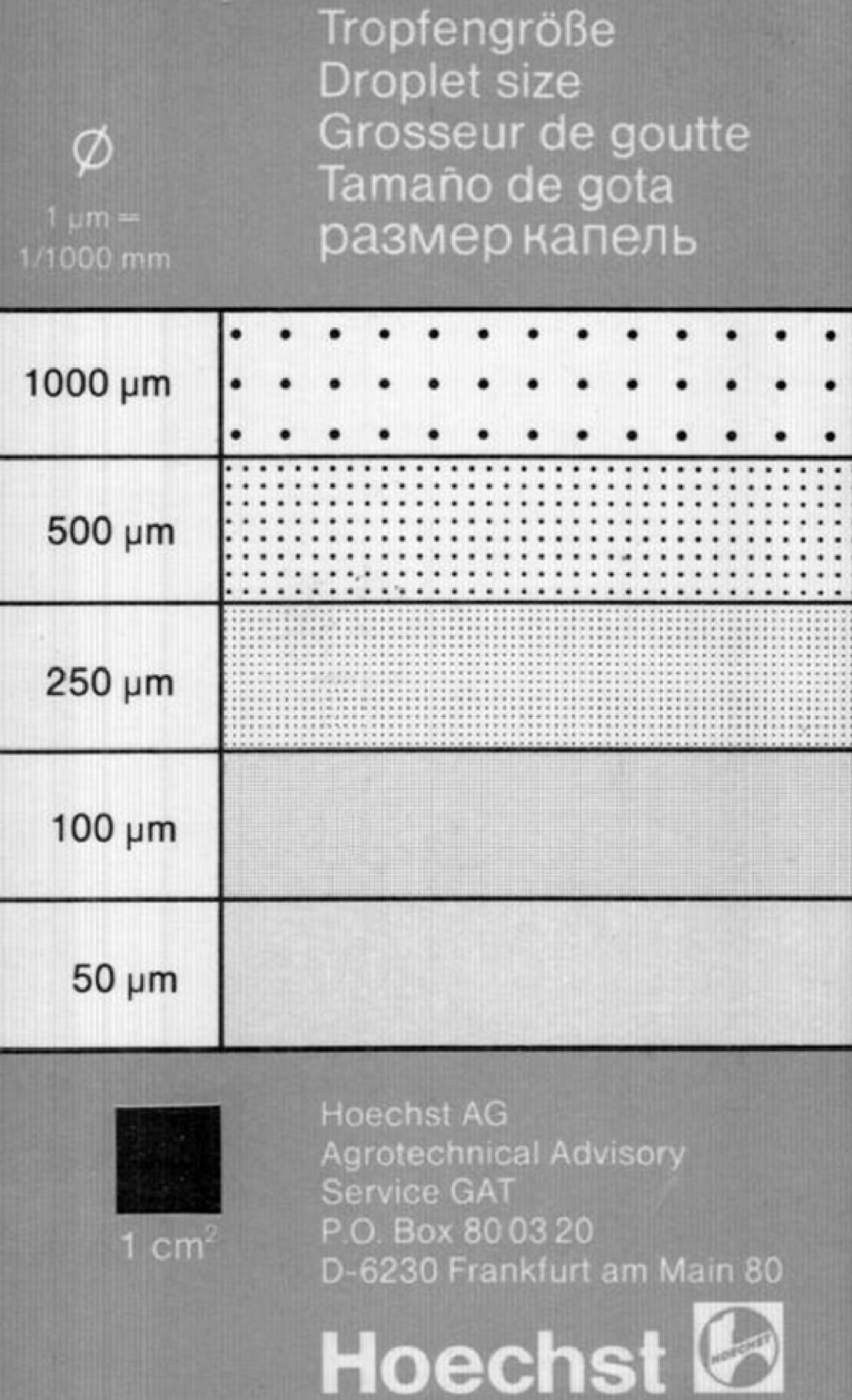}}
   \caption{Control card provided by {\it Hoechst}.}
   \label{fig:controcard}
\end{figure}

\afterpage{

\begin{figure*}[!htb]
   \centering
   \setlength{\fboxsep}{0pt}%
   \begin{subfigure}{.165\textwidth}
     \begin{subfigure}{.5\textwidth}
       \centering
       \fbox{\includegraphics[width=.98\linewidth]{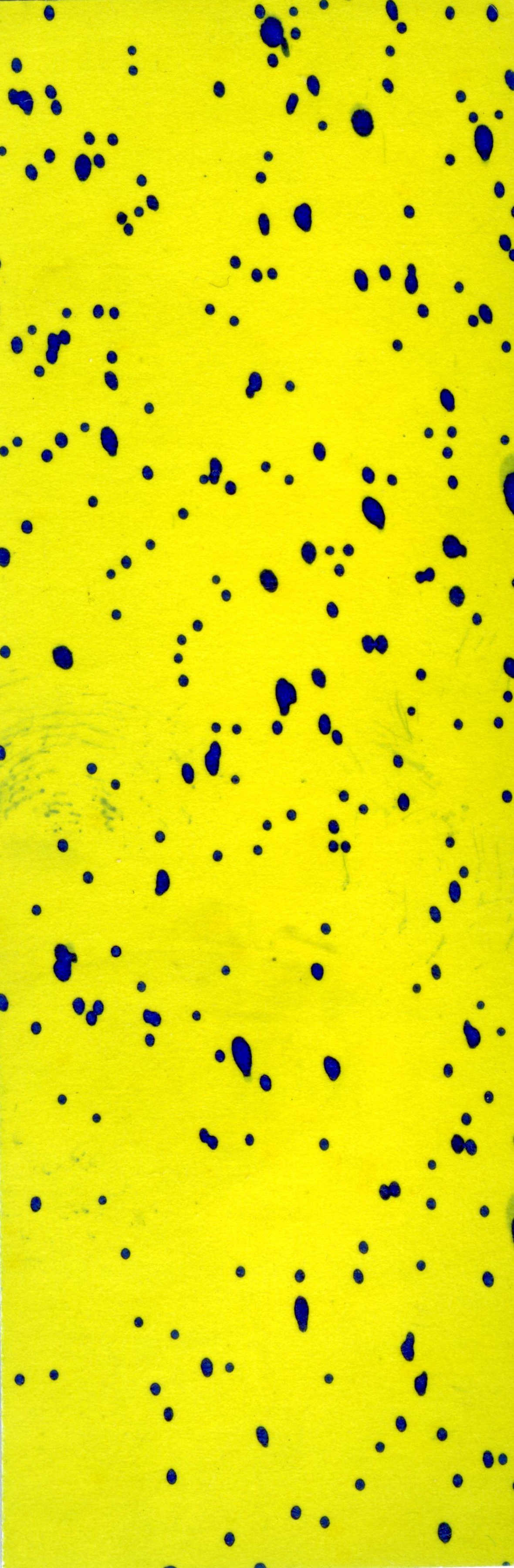}}
     \end{subfigure}%
     \begin{subfigure}{.5\textwidth}
       \centering
       \fbox{\includegraphics[width=.98\linewidth]{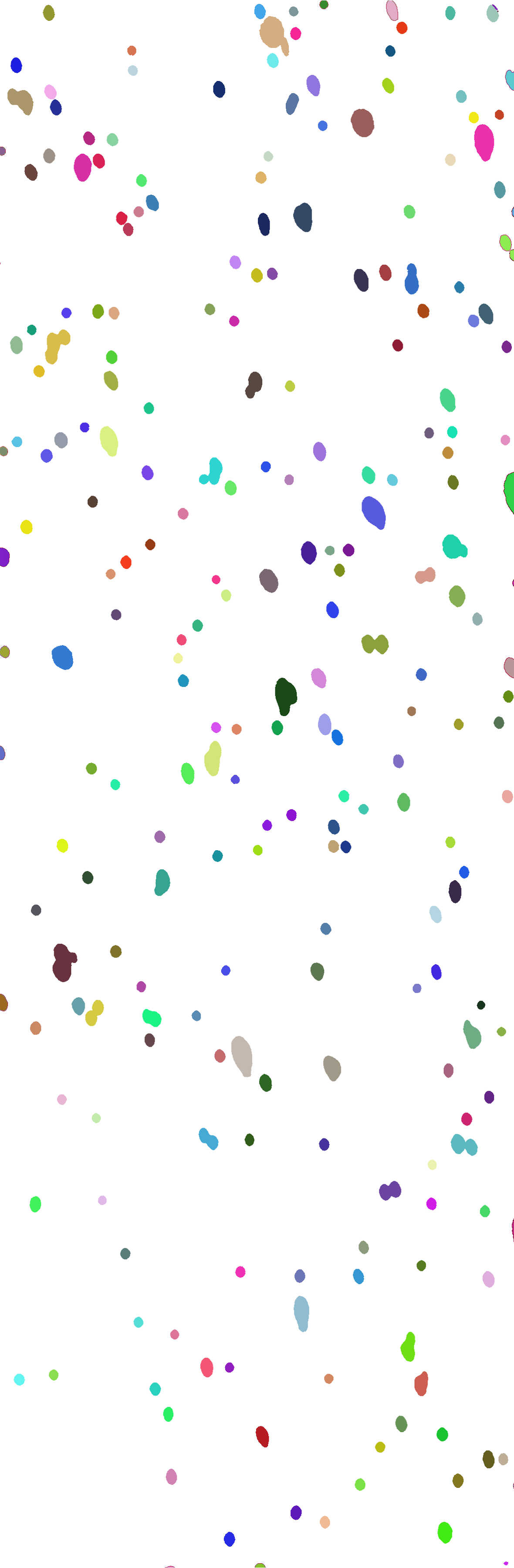}}
     \end{subfigure}%
     \caption{}
     \label{fig:sparse1-7}
   \end{subfigure}
   \hspace{-1.8mm}
   \begin{subfigure}{.165\textwidth}
     \begin{subfigure}{.5\textwidth}
       \centering
       \fbox{\includegraphics[width=.98\linewidth]{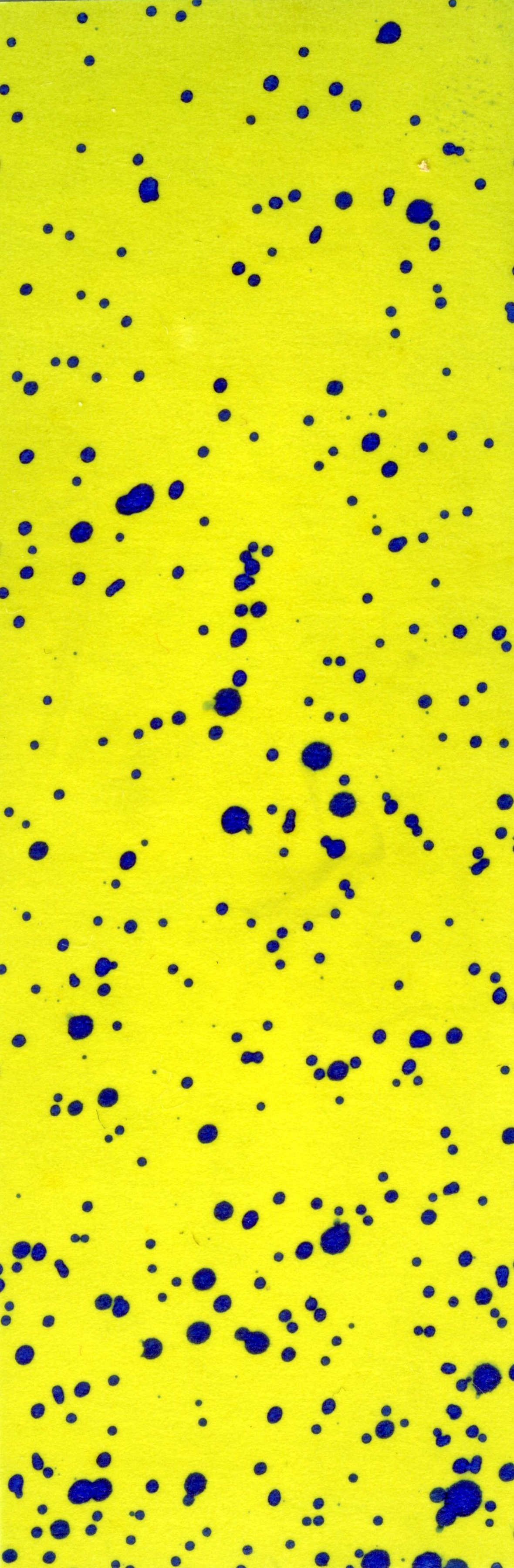}}
     \end{subfigure}%
     \begin{subfigure}{.5\textwidth}
       \centering
       \fbox{\includegraphics[width=.98\linewidth]{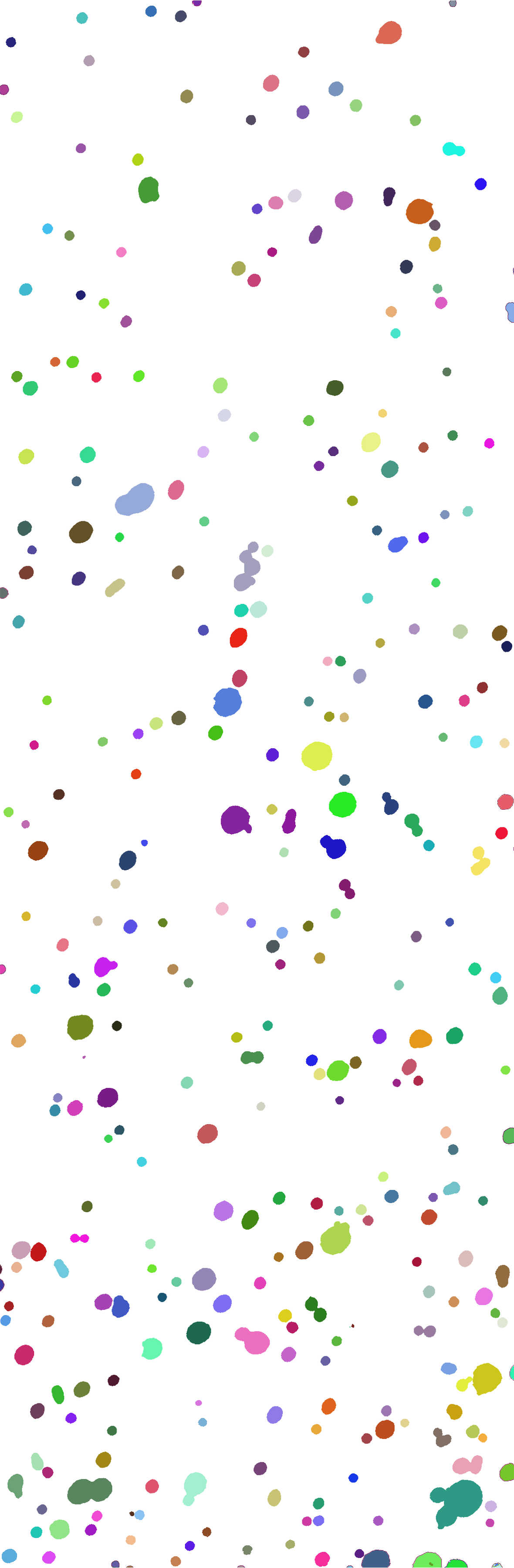}}
     \end{subfigure}%
     \caption{}
     \label{fig:sparse3-6}
   \end{subfigure}%
   \hfill
   \begin{subfigure}{.165\textwidth}
     \begin{subfigure}{.5\textwidth}
       \centering
       \fbox{\includegraphics[width=.98\linewidth]{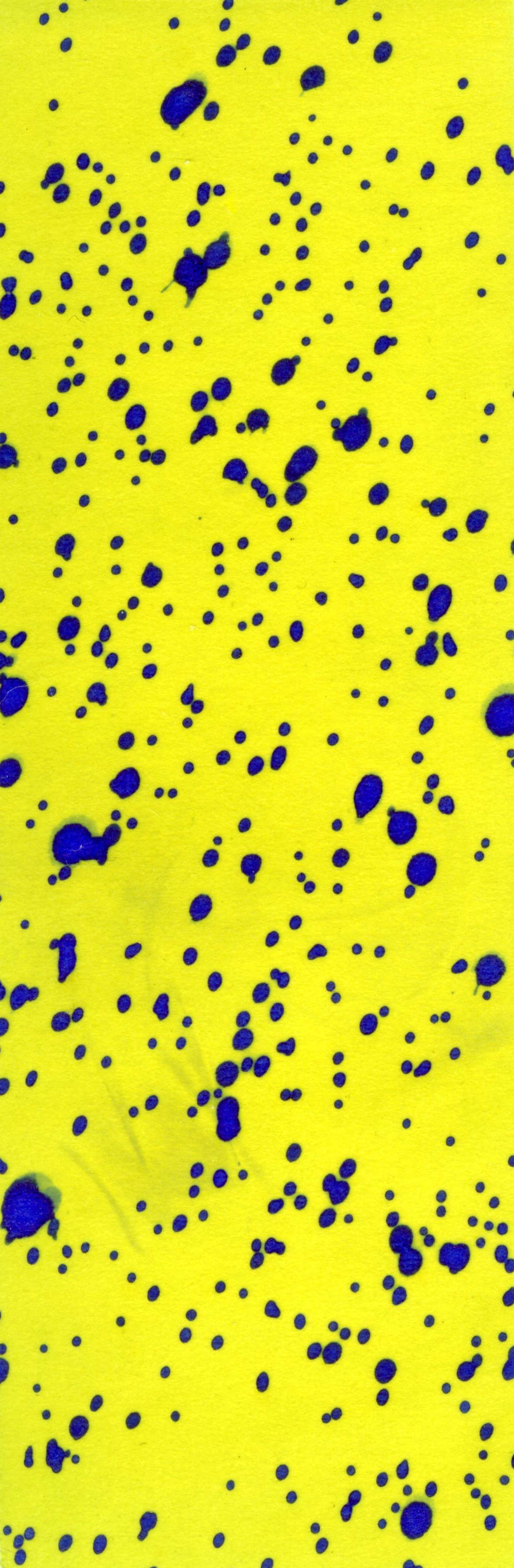}}
     \end{subfigure}%
     \begin{subfigure}{.5\textwidth}
       \centering
       \fbox{\includegraphics[width=.98\linewidth]{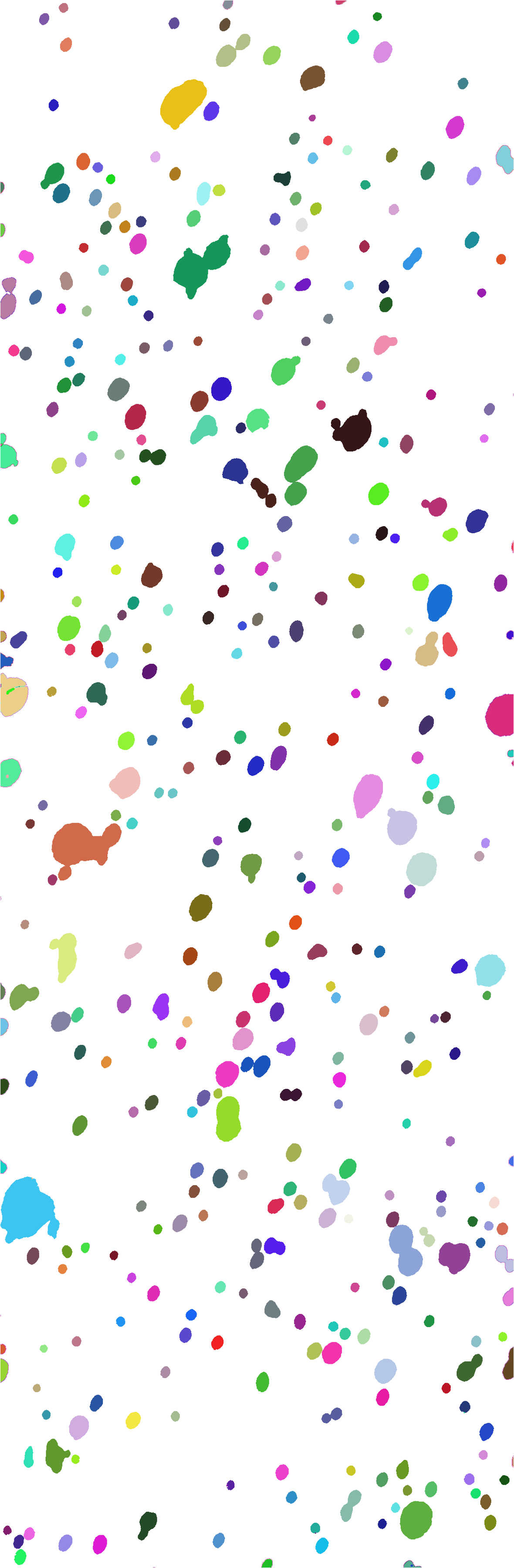}}
     \end{subfigure}%
     \caption{}
     \label{fig:medium1-5}
   \end{subfigure}%
   \hfill
   \begin{subfigure}{.165\textwidth}
     \begin{subfigure}{.5\textwidth}
       \centering
       \fbox{\includegraphics[width=.98\linewidth]{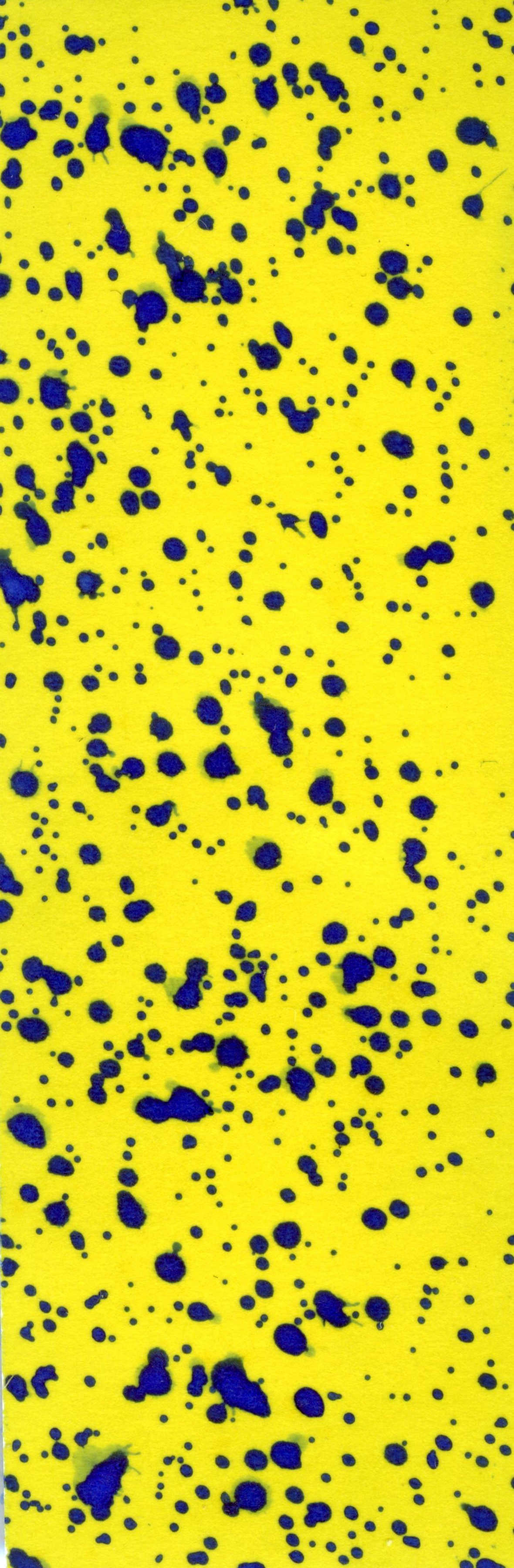}}
     \end{subfigure}%
     \begin{subfigure}{.5\textwidth}
       \centering
       \fbox{\includegraphics[width=.98\linewidth]{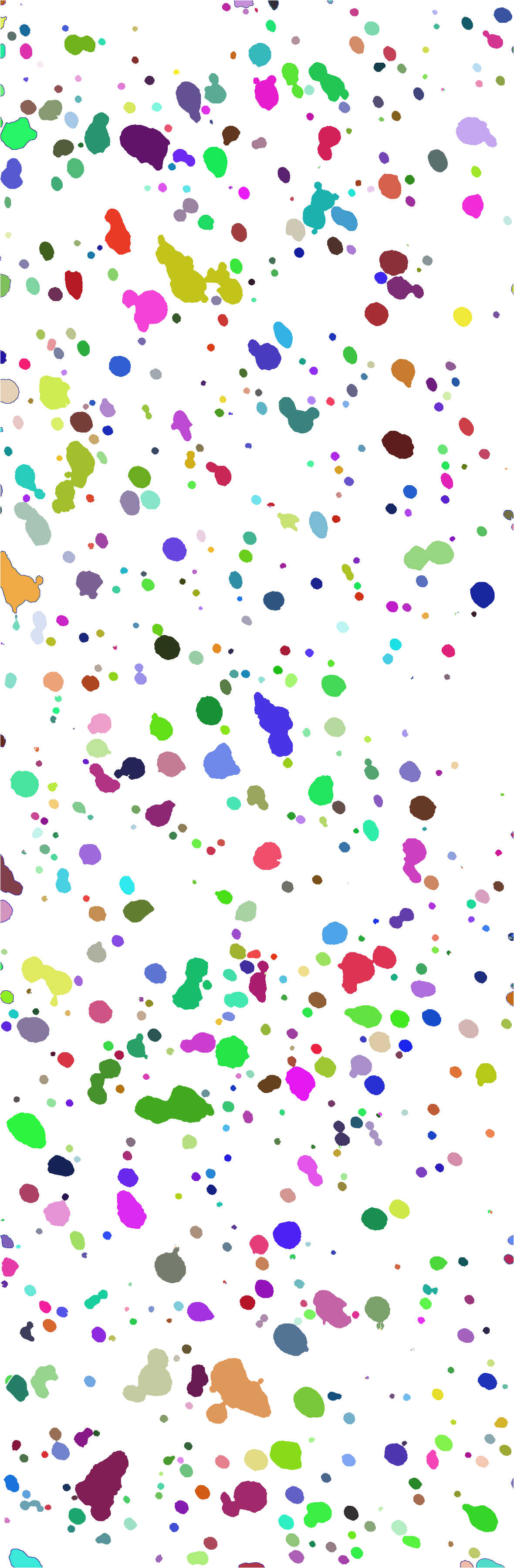}}
     \end{subfigure}%
     \caption{}
     \label{fig:medium3-2}
   \end{subfigure}%
   \hfill
   \begin{subfigure}{.165\textwidth}
     \begin{subfigure}{.5\textwidth}
       \centering
       \fbox{\includegraphics[width=.98\linewidth]{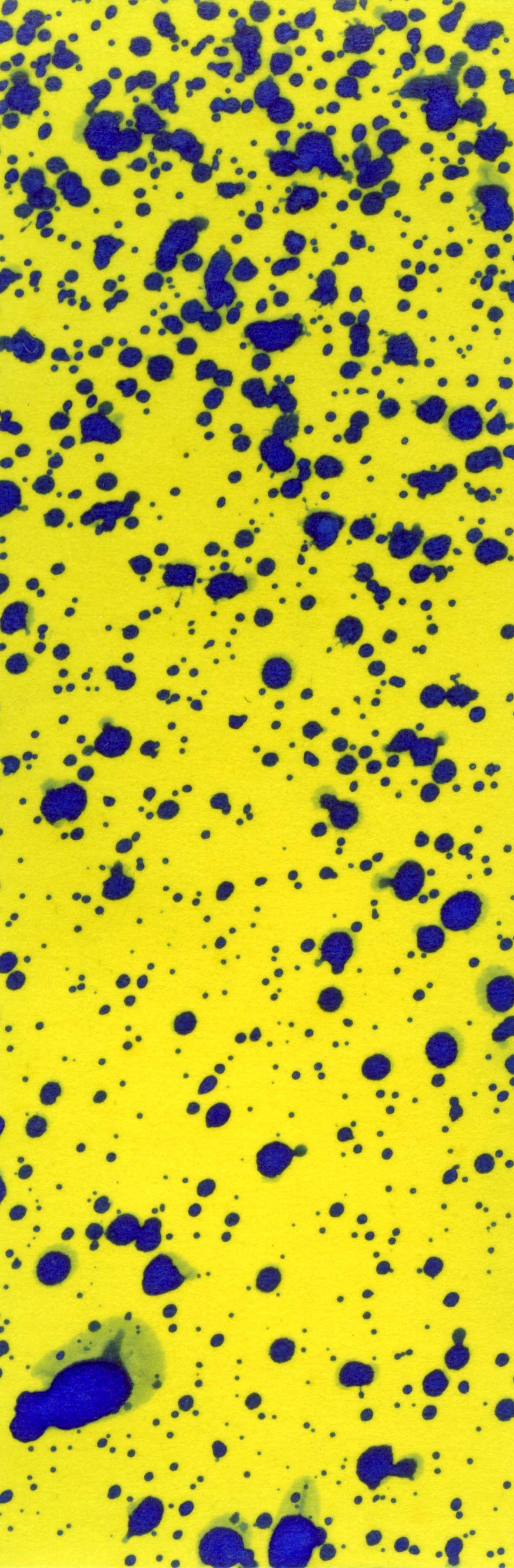}}
     \end{subfigure}%
     \begin{subfigure}{.5\textwidth}
       \centering
       \fbox{\includegraphics[width=.98\linewidth]{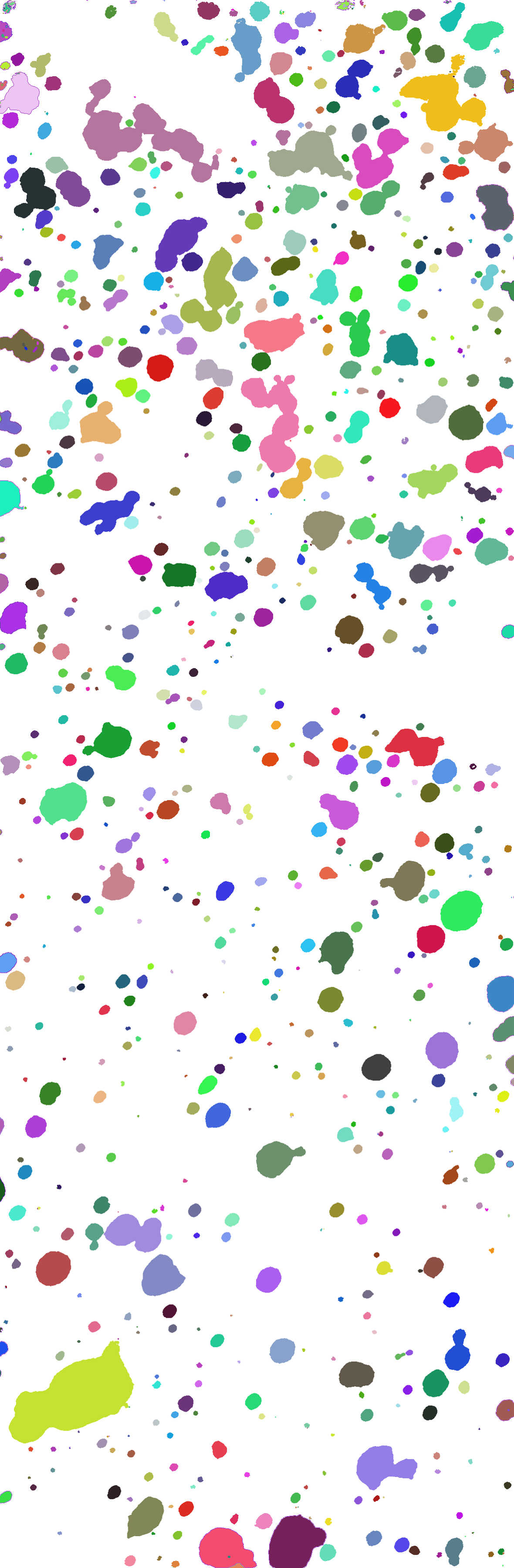}}
     \end{subfigure}%
     \caption{}
     \label{fig:dense1-8}
   \end{subfigure}%
   \hfill
   \begin{subfigure}{.165\textwidth}
     \begin{subfigure}{.5\textwidth}
       \centering
       \fbox{\includegraphics[width=.98\linewidth]{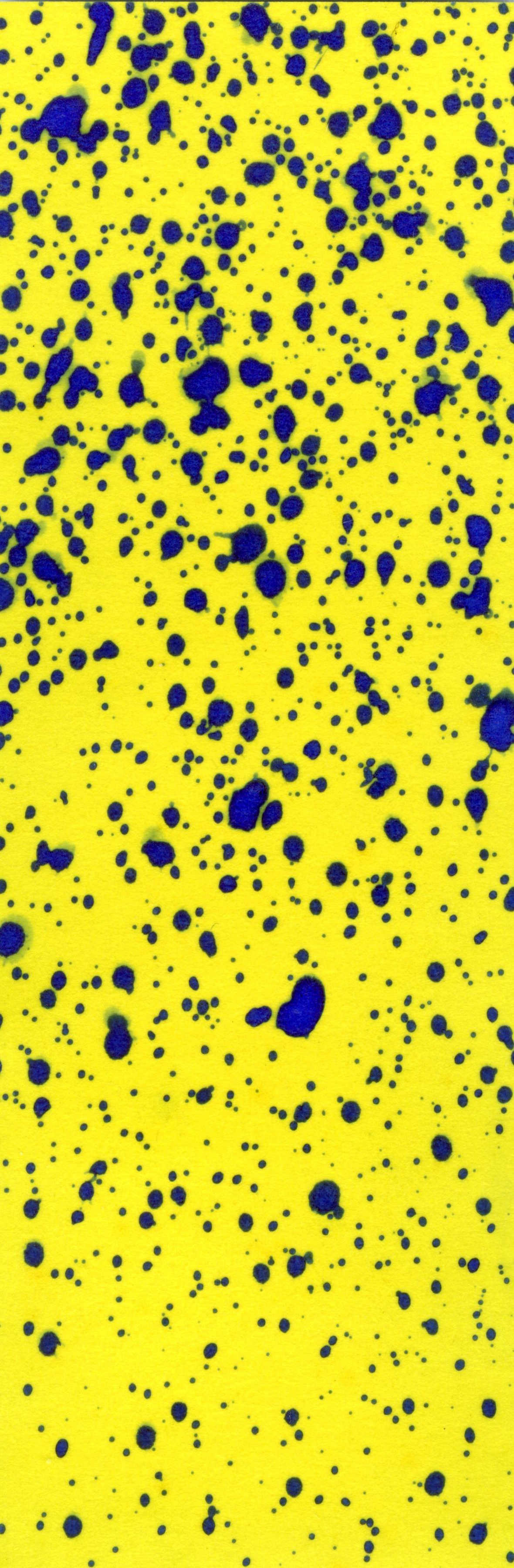}}
     \end{subfigure}%
     \begin{subfigure}{.5\textwidth}
       \centering
       \fbox{\includegraphics[width=.98\linewidth]{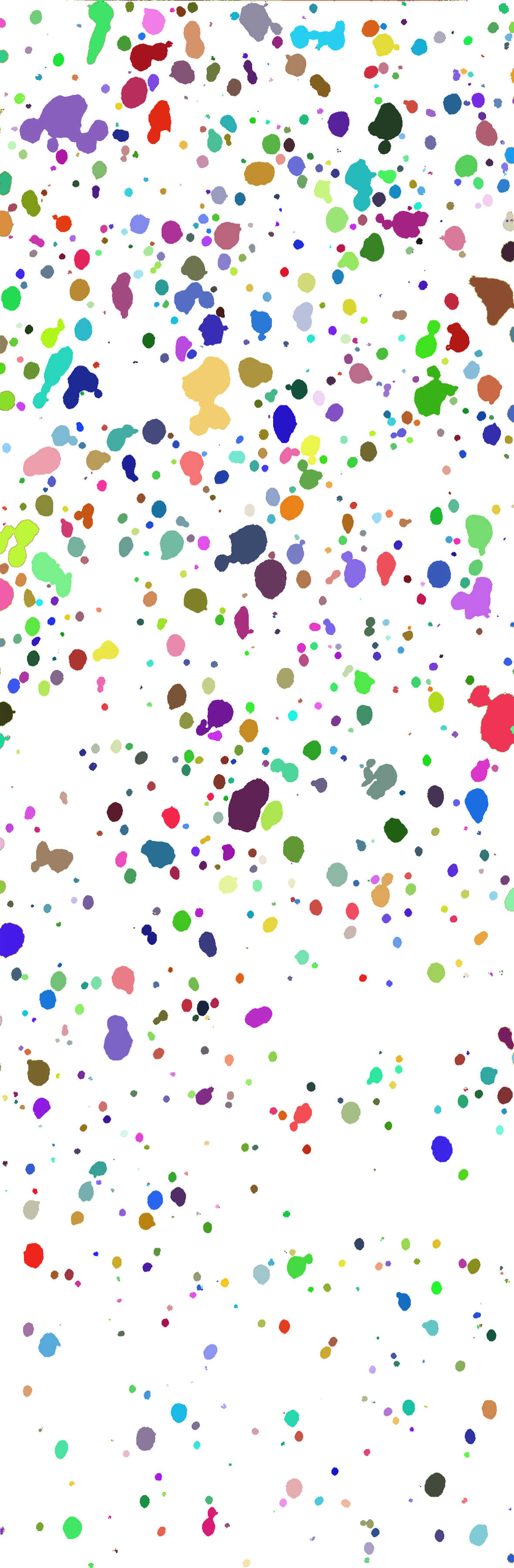}}
     \end{subfigure}%
     \caption{}
     \label{fig:dense2-3}
   \end{subfigure}%
   \caption{Drop identification over cards used in a real crop.
   The cards are categorized in: sparse -- images (a) and (b); medium -- images (c) and (d); and dense -- images (e) and (f).
   The ones on the left are the original cards, and at the right are the segmented cards.}
   \label{fig:realcards}
\end{figure*}

\begin{table*}[!htb]
   \centering
   \scalebox{0.9}{%
   \begin{tabular}{c|c|c|c|c|c|c|c|c|c}
   \hline
   \multirow{2}{*}{\textbf{Diameter ($\mu m$)}} & \multicolumn{2}{c|}{\textbf{DropLeaf}}    &
   \multirow{2}{*}{\textbf{Error}}              & \multicolumn{2}{c|}{\textbf{DepositScan}} &
   \multirow{2}{*}{\textbf{Error}}              & \multicolumn{2}{c|}{\textbf{Microscope}}  & \multirow{2}{*}{\textbf{Error}} \\ \cline{2-3}
                                                & \textbf{Area ($\mu m ^2$)}                &
   \textbf{Diameter ($\mu m$)}                  &                                           &
   \textbf{Area ($\mu m ^2$)}                   & \textbf{Diameter ($\mu m$)}               &
                                                & \textbf{Area ($\mu m ^2$)}                &
   \textbf{Diameter ($\mu m$)}                  & \\ \hline
       50    & 2,693   & 58    & 16\%  & 6,093   & 88  & 76\%  & 3,390   & 66    & 32\%  \\ \hline
       100   & 15,687  & 141   & 41\%  & 21,505  & 165 & 65\%  & 15,906  & 142   & 42\%  \\ \hline
       250   & 53,470  & 246   & 1.6\% & 52,688  & 259 & 3.6\% & 45,342  & 240   & 4\%   \\ \hline
       500   & 214,970 & 467   & 6.6\% & 196,236 & 500 & 0\%   & 201,924 & 507   & 1.4\% \\ \hline
       1,000 & 901,811 & 1,009 & 0.9\% & 777,954 & 995 & 0.5\% & 797,752 & 1,008 & 0.8\% \\ \hline
   \end{tabular}}
   \caption{DropLeaf measures compared to the tool DepositScan and to a stereoscopic microscope.}
   \label{tab:microscope}
\end{table*}

\begin{table*}[!htb]
   \centering
   \scalebox{0.9}{%
   \begin{tabular}{c|c|c|c|c|c|c}
       \hline
       \multicolumn{7}{c}{\textbf{Dropleaf}} \\ \hline
       Sample     & \multirow{2}{*}{Drops} & \multirow{2}{*}{Area ($\mu m^2$)} & Density       & Coverage     & Volumetric Median  & Diameter Relative \\
       Card       &                        &                                   & (drop/$cm^2$) & Density (\%) & Diameter ($\mu m$) & Span ($\mu m$)    \\ \hline
       sparse (a) & 255                    & 250,138                           & 12.90         & 4.54\%       & 452                & 1.22              \\ \hline
       sparse (b) & 359                    & 261,464                           & 18.16         & 6.45\%       & 425                & 1.55              \\ \hline
       medium (c) & 448                    & 355,712                           & 22.67         & 9.99\%       & 448                & 1.83              \\ \hline
       medium (d) & 444                    & 357,005                           & 22.46         & 9.71\%       & 428                & 2.22              \\ \hline
       dense (e)  & 923                    & 364,749                           & 46.71         & 18.22\%      & 246                & 3.75              \\ \hline
       dense (f)  & 1,150                  & 215,090                           & 58.19         & 15.44\%      & 239                & 3.40              \\ \hline
   \end{tabular}}
   \caption{Drop identification over cards used in a real crop.}
   \label{tbl:DropleafResults}
\end{table*}

\begin{table*}[!hbt]
   \centering
   \scalebox{0.9}{%
   \begin{tabular}{c|c|c|c|c|c}
       \hline
       \multicolumn{6}{c}{\textbf{Dropleaf}}                                                                       \\ \hline
       \multicolumn{2}{c|}{Diameter ($\mu m$)}  & Coverage      & Density      & Coverage     & Volumetric Median  \\ \cline{1-2}
       Controlled & Measured & Area ($\mu m^2$) & (drop/$cm^2$) & Density (\%) & Diameter ($\mu m$) \\ \hline
       50         & 58       & 2,693            & 594.31        & 5.32\%       & 58                 \\ \hline
       100        & 141      & 15,687           & 399.01        & 15.01\%      & 141                \\ \hline
       250        & 246      & 53,470           & 229.73        & 23.46\%      & 246                \\ \hline
       500        & 467      & 214,970          & 37.20         & 11.8\%       & 467                \\ \hline
       1,000      & 1,009    & 901,811          & 3.65          & 3.72\%       & 1,009              \\ \hline
   \end{tabular}}
   \caption{Drop identification over the set of control cards.}
   \label{tbl:Results}
\end{table*}
}

\section{Experimental results}
\label{sec:res}

In this section, we evaluate our methodology in the task of measuring the spray coverage deposition on water-sensitive cards.
The goal is to have our technique correctly identifying the spray drops both in terms of density of spraying (percentage of coverage per $cm^2$) and in terms of drop diameter.
As so, the first set of experiments was conducted over a control card provided by enterprise {\it Hoechst}, demonstrating the accuracy in controlled conditions.
The second set of experiments was conducted over a real water-sensitive card that was used on soy crops, demonstrating that the application works even during {\it in situ} conditions.

\subsection{Control-card experiments}

In this set of experiments, we use the card provided by the {\it Agrotechnical Advisory} of the German enterprise {\it Hoechst}.
The card holds synthetic drops with sizes 50$\mu m$, 100$\mu m$, 250$\mu m$, 500$\mu m$, and 1,000$\mu m$, as shown in Figure~\ref{fig:controcard}; this card is used to calibrate equipment and to assess the accuracy of manual and automatic measuring techniques.
Since the number and sizes of drops are known, this first experiment works as a controlled validation of the methodology.

To measure the drops of the control card, we used a smartphone to capture the image of the card.
In Table~\ref{tbl:Results}, we present the average diameter of the drops, the area covered by the drops given in $cm^2$, the density given in drops per $cm^2$, the coverage density given in percentage of the card area, and the volumetric median diameter.
We do not present the diameter relative span because, as all the drops are equal, there is no significant span.
From the table, it is possible to conclude that the accuracy of the methodology is in accordance with the controlled protocol; that is, the known and measured diameters matched in most of the cases.
Notice that it is not possible to achieve a perfect identification because of printing imperfections and numerical issues that inevitable rise at the micrometer scale.
For example, for 1,000 $\mu m$ drops, the average diameter was 1,007 $\mu m$.
This first validation was necessary to test the ability of the tool in telling apart card background and drops.

For a comparative perspective, in Table~\ref{tab:microscope}, we compare the covered area and the average diameter measured by our tool, by the tool DepositScan, and by a stereoscopic microscope (provided in the work of Zhu {\it et al.}~\cite{Zhu2011}).
The results demonstrated that the stereoscopic microscope had the best performance, as expected, since it is a fine-detail laborious inspection.
DropLeaf presented the best results after the microscope, beating the precision of DepositScan for all the drop sizes, but 500 $\mu m$; for 1,000 $\mu m$ drops, the two tools had a similar performance, diverging in less than $1\%$.
In the experiments, one can notice that the bigger the drop, the smaller the error, which ranged from 41\% to less that 1\%.
For bigger drops, the drop identification is next to perfect; for smaller ones, the error is much bigger; this is because of the size scale.
When measuring drops as small as 50$\mu m$, a single extra pixel detected by the camera is enough to produce a big error.
This problem was also observed in the work of Zhu {\it et al.}~\cite{Zhu2011}).

By analyzing the data, we concluded that the error due to the size scale is predictable.
Since it varies with the drop size, it is not linear; nevertheless, it is a pattern that can be corrected with the following general equation:

\begin{equation}
diameter' = a*diameter^b
\end{equation}

In the case of our tool, we used $a=0.2192733$ and $b=1.227941$.
These values shall vary from method to method, as we observed for DepositScan and for the stereoscopic microscope.

\subsection{Card experiments}
\label{sec:production}

In the second set of experiments, we used six cards evaluated in the work of Cunha {\it et al.}~\cite{cunhaBE2012}.
The cards were separated into three groups of two cards, which were classified as sparse, medium, and dense with respect to the density of drops, as can be verified in Figure~\ref{fig:realcards}.
These experiments aimed at testing the robustness of the methodology, that is, its ability in identifying drops even when they are irregular and/or they have touching borders.
Table~\ref{tbl:DropleafResults} shows the numerical results, including the number of drops, the coverage area, the density, the coverage density, the volumetric median diameter, and the diameter relative span.
In this case, the table must be interpreted along with the figure, which presents the drops as identified by our methodology.
The first four measures can be inspected visually.
It is also possible to see that the right-hand side images in Figure~\ref{fig:realcards} (the tool's results stressed with colored drops), demonstrate that the segmentation matches the expectations of a quick visual inspection.
The drops at the left are perfectly presented on the right.
Other features are also noticeable.
Density, for instance, raises as we visually inspect Figure~\ref{fig:realcards}(a) to Figure~\ref{fig:realcards}(f); the corresponding numbers in the table raise similarly.
Counting the number of drops requires close attention and a lot of time; for the less dense Figure~\ref{fig:realcards}(a) and Figure~\ref{fig:realcards}(b), however, it is possible to verify the accuracy of the counting and segmentation provided by the tool.

The last two measures, VMD and DRS, provide parameters to understand the distribution of the drops' diameters.
For example, it is possible to see that, being more dense, cards (e) and (f) had a smaller median and a larger span of diameters.
These measures indicate that the spraying is irregular and that it needs to be adjusted.
Meanwhile, cards (a) and (b) are more regular, but not as dense as necessary, with a lot of blank spots.
Cards (c) and (d), in turn, have a more uniform spraying and a more regular coverage.

\section{Drop detection issues}
\label{sec:discussao}

This section discusses issues to be considered when developing technologies for spray card inspection.
We faced such issues during our research; we discuss such issues as a further contribution that shall guide other researchers that deal with the same and with related problems.

\subsection{Coverage factor}
\label{subs:coverage}

In our experience, we noticed that when the spraying gets too dense, not all of the information about the drops can be detected, no matter which technique is used for measuring; for instance, information about the number of drops, and their diameter distribution cannot be tracked anymore.
This effect was already pointed out by Fox {\it et al.}~\cite{Fox2003}, who claims that a total coverage on the card above 20\% causes results to be unreliable; and coverages close to 70\% are unfeasible.

This is because, with too much spray, the drops fall too close one to each other, causing overlaps; visually, is it like two or more drops became one.
Effectively, this is what happens in the crop due to the intermolecular forces present in the water drops, which causes them to merge, forming bigger drops.
Hence, it is needed caution, no matter which technique of assessment, whenever the total coverage area surpasses 20\%, a circumstance when the diameter distribution is no longer accurate, and one must rely only on the coverage area for decision making.
Although the diameter is not available, the large drops that might be detected indicate an excessive amount of pesticide or a malfunctioning of the spray device.

\subsection{Angle of image capture}
\label{subs:angle}

We also noticed that the image processing methodology used to detect the drops of all the studies presented so far, including ours, works only if the capture angle of the card is equal to 90 degrees.
That is, the viewing angle of the camera/scanner must be orthogonal to the spray card surface.
This is necessary because the pixels of the image are converted into a real-world dimension to express the diameter of the drops in $\mu m$; therefore, it is necessary that the dimensions of the image be homogeneous with respect to scale.
In case, the capture angle is not of 90 degrees, the image is distorted, resulting in different scales in each part of the image.
For flatbed scanners, this is straightforward to guarantee; however, for handheld devices (cameras and smartphones), additional care is necessary.
In such cases, one might need a special protocol in order to capture the image, like using a tripod, or some sort of apparatus to properly place the capturing device with respect to the spray card.
This problem might also be solved by means of an image processing algorithm to remove eventual distortions, in which case, additional research and experimentation are necessary.

\subsection{Minimum dots per inch (dpi)}
\label{subs:dpi}

Our experiments also reviewed that there must be a minimum amount of information on the spray card images in order to achieve the desired precision regarding the drops' diameter.
This minimum information is expressed by the {\it dots per inch} (dpi) property of the image capturing process; dpi is a well-known resolution measure that expresses how much pixels are necessary to reproduce the real-world dimension of one linear inch.
If not enough pixels are captured per inch of the spray card during the capturing process, it becomes impossible to estimate, or even to detect, the diameter of the smallest drops.
This might influence the diameter distribution analysis hiding problems in the spraying process.

In order to guide our research and development, we calculated and tested on the minimum dpi's that are necessary for each desired drop diameter.
In Table~\ref{tab:dpis} one can see the minimum number of pixels to express each drop diameter for each dpi value; notice that some cells of the table are empty (filled with a hyphen) indicating that the diameter cannot be computationally expressed in that dpi resolution.
Also, notice that, in the columns, the number of pixels for one same diameter increases with the resolution.
Obviously, the more information, the more precision at the cost of more processing power, substantially more storage, and more network bandwidth when transferring images.
From the table, it is possible to conclude that 600 dpi is the minimum resolution for robust analyses, since it can represent diameters as small as 50 $\mu m$; meanwhile, a resolution of 1,200 dpi, although even more robust, might lead to drawbacks regarding the management of image files that are way too big.
Notwithstanding, the fact that a resolution is enough to represent a given diameter is not a guarantee that drops with that diameter size will be detected; this is because the detection depends on other factors such as the quality of the lenses, and the image processing algorithm.

Table~\ref{tab:dpis} is a guide for developers willing to computationally analyze spray cards, and also for agronomists who are deciding which equipment to buy in face of their needs.
\begin{table}[!htb]
   \centering
   \begin{tabular}{c|c|c|c|c|c|c|c}
       \hline
       \backslashbox{$\mu m$\kern-1em}{\kern-1em dpi} & 50 & 100 & 300 & 600 & 1200 & 2400 & 2600 \\ \hline
       10     & -  & -   & -   & -   & -    & -    & 1    \\ \hline
       50     & -  & -   & -   & 1   & 2    & 5    & 5    \\ \hline
       100    & -  & -   & 1   & 2   & 5    & 9    & 10   \\ \hline
       250    & -  & 1   & 3   & 6   & 12   & 24   & 26   \\ \hline
       500    & 1  & 2   & 6   & 12  & 24   & 47   & 51   \\ \hline
       1,000  & 2  & 4   & 12  & 24  & 47   & 94   & 102  \\ \hline
       10,000 & 20 & 39  & 118 & 236 & 472  & 945  & 1024 \\ \hline
   \end{tabular}
   \caption{Pixels needed to represent a given length, given a dpi.}
   \label{tab:dpis}
    \vspace{-.75cm}
\end{table}
\section{Conclusions}
\label{sec:conclusao}

We introduced DropLeaf, a portable application to measure the pesticide coverage by means of the digitalization of water-sensitive spray cards.
We verified that the precision of DropLeaf was enough to allow the use of mobile phones as substitutes for more expensive and troublesome methods of quantifying pesticide application in crops.
The methodology was instantiated in a tool to be used in the inspection of real-world crops.
We tested our tool with two datasets of water-sensitive cards; our experiments demonstrated that DropLeaf accurately tracks drops, being able to measure the pesticide coverage and the diameter of the drops.
Furthermore, our mobile application detects overlapping drops, an important achievement because a finer precision provides not only better accuracy but also more information.

\vfill
\pagebreak
\section*{Acknowledgements}

The authors are thankful to Dr. Claudia Carvalho who kindly provided annotated cards; to the Sao Paulo Research Foundation (grants 2011/02918-0, 2016/02557-0 and CEPID CeMEAI grant 2013/07375-0); to the National Council for Scientific and Technological Development (CNPq); and to the Coordination for the Improvement of Higher Education Personnel (Capes).